\documentclass[conference]{IEEEtran}
\usepackage{times}

\usepackage[numbers]{natbib}
\usepackage{multicol}
\usepackage{amsmath}
\usepackage{amsfonts}
\usepackage{graphicx}
\usepackage{subcaption}
\usepackage[bookmarks=true]{hyperref}
\usepackage{booktabs}
\usepackage{makecell}

\begin{document}

\title{Reinforcement Learning based 6-DoF Maneuvers for Microgravity Intravehicular Docking: \\A Simulation Study with Int-Ball2 in ISS-JEM}


\author{\authorblockN{Aman Arora}
\authorblockA{Space Robotics Research Group\\University of Luxembourg\\aman.arora@uni.lu}
\and
\authorblockN{Matteo El-Hariry}
\authorblockA{Space Robotics Research Group\\University of Luxembourg\\matteo.elhariry@uni.lu}
\and
\authorblockN{Miguel Olivares-Mendez}
\authorblockA{Space Robotics Research Group\\University of Luxembourg\\miguel.olivaresmendez@uni.lu}}


%

\maketitle

\begin{abstract}
Autonomous free-flyers play a critical role in intravehicular tasks aboard the International Space Station (ISS), where their precise docking under sensing noise, small actuation mismatches, and environmental variability remains a nontrivial challenge. This work presents a reinforcement learning (RL) framework for six-degree-of-freedom (6-DoF) docking of JAXA's Int-Ball2 robot inside a high-fidelity Isaac Sim model of the Japanese Experiment Module (JEM). Using Proximal Policy Optimization (PPO)~\cite{schulman2017proximal}, we train and evaluate controllers under domain-randomized dynamics and bounded observation noise, while explicitly modeling propeller drag-torque effects and polarity structure. This enables a controlled study of how Int-Ball2's propulsion physics influence RL-based docking performance in constrained microgravity interiors. The learned policy achieves stable and reliable docking across varied conditions and lays the groundwork for future extensions pertaining to Int-Ball2 in collision-aware navigation, safe RL, propulsion-accurate sim-to-real transfer, and vision-based end-to-end docking.
\end{abstract}

\IEEEpeerreviewmaketitle

\section{Introduction}

Autonomous free-flying robots play a key role in intravehicular operations aboard crewed spacecraft such as the ISS. Platforms like SPHERES~\cite{mohan2009spheres}, Astrobee~\cite{smith2016astrobee}, CIMON~\cite{schmitz2020towards}, and JAXA's Int-Ball series~\cite{mitani2019int} support inspection, monitoring, and crew assistance. A critical capability for such platforms is autonomous navigation and docking~\cite{carlino2024lessons, yamaguchi2024int}, enabling unassisted traversal of cluttered interiors and return to charging stations.

Microgravity environments such as the ISS JEM~\cite{mitani2019int, yamaguchi2024int} impose strict requirements: 6-DoF control without GPS, operation in confined spaces, and robustness to disturbances and sensor noise. Classical control methods (e.g. LQR, MPC), although reliable in nominal conditions, often struggle to generalize under uncertainties arising from imperfect state estimation, unmodeled propulsion effects, and variable contact-free dynamics ~\cite{tipaldi2022reinforcement}, and the recent systems like Astrobee \cite{smith2016astrobee} and Int-Ball \cite{mitani2019int} still rely heavily on pre-programmed control logic.

Reinforcement learning (RL) offers a data-driven alternative for spacecraft guidance, with prior work showing policy resilience to model mismatch, actuator failure, and uncooperative targets \cite{hovell2021deep}. Yet, most prior work focuses on open-space rendezvous or planar docking, leaving several practical challenges unexplored: (i) actuation using multi-propeller platforms rather than thrusters, (ii) tightly constrained intravehicular environments such as the ISS JEM module, and (iii) the physical asymmetries introduced by aerodynamic drag, fan polarity, and geometric actuation coupling. While most existing research focuses on perception and state estimation for free-flyers, learning-based 6-DoF control in confined intravehicular spaces has remained 
largely unexplored \cite{kang2024astrobee}.

This work presents a PPO-trained 6-DoF docking controller for JAXA's Int-Ball2, evaluated in a high-fidelity Isaac Lab simulation of the JEM. Our hypothesis is that an attempt to accurately model the Int-Ball2 propeller drag-torque dynamics and polarity structure materially improves docking stability and final alignment, and that RL can leverage this additional torque channel to achieve robust performance under uncertainty.

\begin{figure}[t]
  \centering
  \includegraphics[width=\columnwidth]{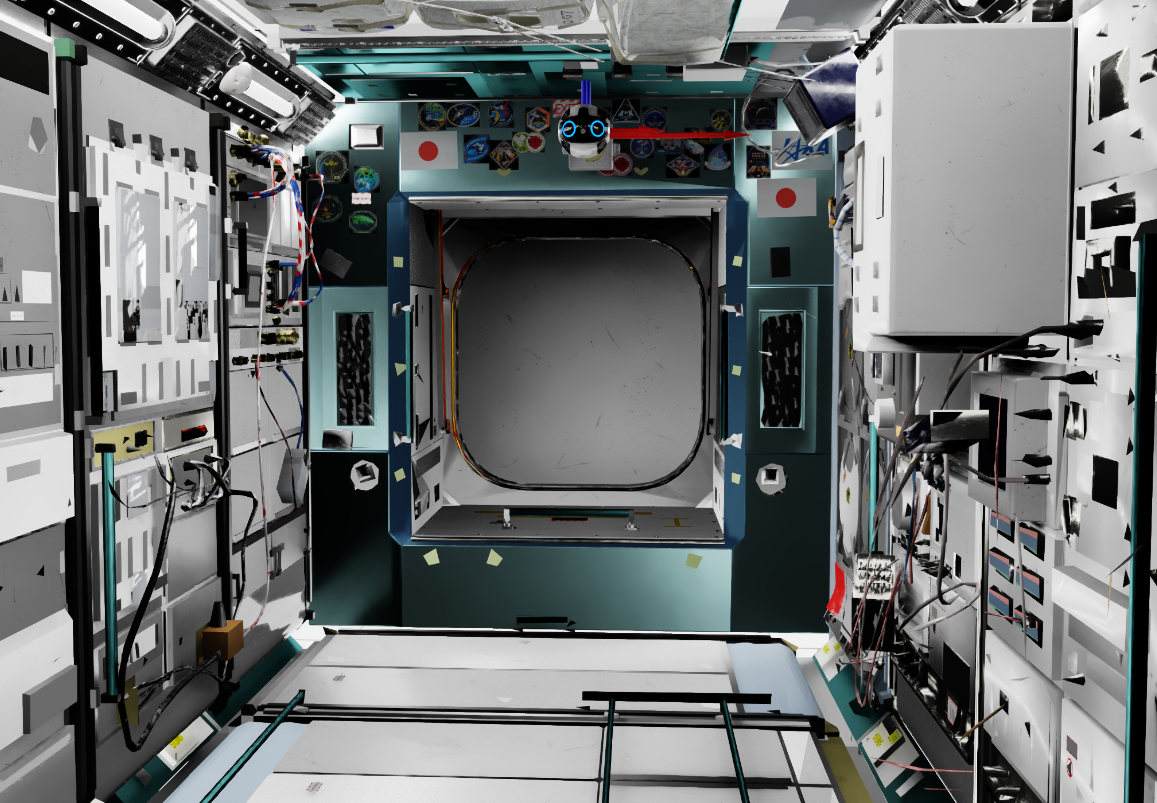}
  \caption{Isaac Lab simulation of Int-Ball2 performing a 6‑DoF
           docking maneuver inside the ISS-JEM.}
  \label{fig:sim_overview}
  \vspace{-6pt} 
\end{figure}

To test this, we incorporate task- and robot-level reward shaping, including a physically motivated drag-torque penalty~\cite{Mitani2023IntBall2}, safe-region-aware navigation, and domain-randomized dynamics. The resulting framework enables systematic ablations on drag dynamics, polarity structure, and regularization, and establishes a foundation for future research  extensions pertaining to learned-controller development for Int-Ball2.

\section{Method}

\subsection{Problem Formulation.} We formulate the docking approach task as a 6-DoF goal-reaching problem, where the Int-Ball2 robot must align its pose with that of a fixed docking port inside the ISS Kibo module. This is modeled as a Markov Decision Process $(\mathcal{S}, \mathcal{A}, \mathcal{P}, r, \gamma)$ and trained using actor-critic PPO \cite{schulman2017proximal} within our modular simulation framework \texttt{RoboRAN} \cite{elroboran}, built on top of NVIDIA Isaac Lab. Robots and tasks are configured independently, allowing flexible robot-task pairing.

\subsection{\textit{DockToStation} Environment}

\smallskip
\noindent
\textbf{Observation, Action Spaces and Rewards}

The agent has access to the complete state information available in the simulation. The observed quantities are described in Table \ref{tab:obs}. The observation vector consists of calculated errors from the goal, and the velocities in the agent's body frame, as well as the policy's continuous thrust command for the respective propeller, from the preceding timestep. The continuous action space is defined as $\mathcal{A} = [-1,1]^{N_{\mathrm{thr}}}$, representing normalized thrust commands for each of the eight propellers, which are linearly mapped to $[0,1]$ prior to actuation. The rewards used for the \textit{DockToStation} task are described in Table \ref{tab:rewards}.

\begin{table*}[t]
\centering
\caption{Observation structure for the \textit{DockToStation} task 
($\dim(o_t)=23$; 15 task-specific state variables $+$ 8 thruster commands).
All quantities are expressed in the robot body frame. 
$\Delta\mathbf{p}$ and $\Delta\mathbf{q}$ denote the position and orientation errors,
$\mathbf{v}_{\mathrm{lin}}$, $\mathbf{v}_{\mathrm{ang}}$ are body-frame velocities,
and $\mathbf{u}_{t}$ represents the applied thruster command vector.}
\label{tab:obs}
\renewcommand{\arraystretch}{1.2}
\begin{tabular}{c p{3.5cm} p{9.5cm}}
\hline
\textbf{Dim} & \textbf{Component} & \textbf{Included Variables} \\
\hline
3  & Position error & $\Delta\mathbf{p} = [\,\Delta p_x,\ \Delta p_y,\ \Delta p_z\,]$ \\[2pt]
6  & Orientation error (6D) & $\mathrm{Rot6D}(\Delta\mathbf{q}) =$ first two columns of the rotation matrix 
derived from $\mathbf{q}_{\mathrm{cur}}^{-1}\!\otimes\!\mathbf{q}_{\mathrm{tgt}}$ \\[2pt]
3  & Linear velocity & $\mathbf{v}_{\mathrm{lin}} = [\,v_x,\ v_y,\ v_z\,]$ \\[2pt]
3  & Angular velocity & $\mathbf{v}_{\mathrm{ang}} = [\,\omega_x,\ \omega_y,\ \omega_z\,]$ \\[2pt]
8  & Thruster commands & $\mathbf{u}_{t-1} = [\,u_{1,t-1},\ldots,u_{8,t-1}\,]$ \\[2pt]
\hline
\end{tabular}
\end{table*}

The Int-Ball2 is modeled as a rigid body with eight thrusters at positions $\mathbf{r}_i$ and unit thrust directions $\mathbf{n}_i$. For a normalized propeller command $u_{i,t}\!\in[0,1]$, the aerodynamic force and resulting body torque are
\begin{align}
\mathbf{f}_i &= -u_{i,t}\,f_{\max,i}\,\mathbf{n}_i, \\
\boldsymbol{\tau}_i &= \mathbf{r}_i \times \mathbf{f}_i,
\end{align}
where $f_{\max,i}$ includes the per-propeller thrust-scaling factors provided in the Int-Ball2 propulsion characterization~\cite{Mitani2023IntBall2}. As demonstrated experimentally in \cite{Mitani2023IntBall2}, each propeller produces a small reaction torque due to blade-tip aerodynamic drag. The propulsion system intentionally alternates CW/CCW spin polarity so that, when operated symmetrically, these drag torques cancel. Unmodeled or uncompensated drag torque leads to significant yaw drift and undesirable cumulative attitude errors, and is therefore essential to study in both training and simulation.

\paragraph{Penalty for residual drag torque.}
Let $s_j\!\in\{-1,+1\}$ denote the spin polarity of the $j$-th propeller, and
$D_j$ its thrust magnitude after scaling.  
Following~\cite{Mitani2023IntBall2}, the simplified net residual drag torque is modeled as
\begin{align}
\tau_{\mathrm{drag}}
=
\sum_{j=1}^{N} s_j\,k_{\mathrm{drag}}\,|D_j|,
\end{align}
where $k_{\mathrm{drag}}$ is an empirical coefficient derived from
Int-Ball2 torque-response measurements. We have taken the coefficient for 1 propeller from \cite{Mitani2023IntBall2}, and used the same for all propellers, to calculate the reward $R_{\mathrm{drag}}$, in Table \ref{tab:rewards}, for this study.

\paragraph{Drag-torque dynamics in simulation.}
We inject the same simplified model into the dynamics: in addition to the
standard thrust-induced torque $\mathbf{r}_j \times \mathbf{f}_j$, each
propeller contributes a drag-torque vector
\begin{align}
\boldsymbol{\tau}^{\mathrm{world}}_{\!j}
=
s_j\,k_{\mathrm{drag}}\,|D_j|\,\mathbf{a}_j,
\end{align}
where $\mathbf{a}_j$ is the rotor spin axis expressed in the world frame. The total torque applied at each propeller is the sum of these two terms, enabling controlled ablations on (i) the presence of drag-torque dynamics, (ii) polarity structure, and (iii) the drag-penalty term during learning.

\begin{table*}[t]
\centering
\caption{Reward components used for training in the \textit{DockToStation} task.
The total reward is computed as a weighted sum of all terms $R_i$, each scaled by a weight in the task configuration.}
\label{tab:rewards}
\renewcommand{\arraystretch}{1.2}
\begin{tabular}{l p{5.6cm} p{9.3cm}}
\hline
\textbf{Reward Term} & \textbf{Description} & \textbf{Formulation} \\ 
\hline
$R_{\text{pose}}$ 
& Exponential shaping on position and orientation errors ($e_\theta$). Encourages precise 6-DoF alignment with the docking port pose.
& $e^{-\,\|\Delta\mathbf{p}\|_2 / \kappa_p}
 + e^{-\,e_\theta / \kappa_o}$ \\[3pt]

$R_{\mathrm{vel}}$
& Encourages bounded linear and angular speeds within nominal limits.
& $\left[\|\mathbf{v}_{\mathrm{lin}}\|_2 - v_{\min}\right]_{0}^{v_{\max}-v_{\min}}
 + \left[\|\mathbf{v}_{\mathrm{ang}}\|_2 - \omega_{\min}\right]_{0}^{\omega_{\max}-\omega_{\min}}$ \\[3pt]

$R_{\mathrm{boundary}}$
& Penalizes deviation from the operational region around the docking port through an exponential boundary envelope ($d_b$ being the boundary distance).
& $e^{-\,d_b/\kappa_b}$ \\[3pt]

$R_{\mathrm{prog}}$
& Rewards forward progress toward the docking goal between successive timesteps.
& $(d_{\mathrm{prev}} - d)\,(d_{\max} - d)$ \\[3pt]

$R_{\mathrm{cuboid}}$
& Magnitude-proportional penalty applied when the agent exits the defined safe docking envelope ($\delta_i$ being the cuboid violation offsets).
& $-\sqrt{\sum_i \delta_i^2}$ \\[3pt]

$R_{\mathrm{drag}}$
& Penalizes residual torque imbalance arising from CW/CCW drag-torque polarity mismatch.
& $-\,\big|\sum_j s_j k_{\mathrm{drag}} D_j \big|$ \\[3pt]

$R_{\mathrm{act}}$
& Action penalty penalizing abrupt variations in thruster commands to encourage smooth control transitions.
& $\|\mathbf{u}_t - \mathbf{u}_{t-1}\|_2^2$ \\[3pt]

$R_{\mathrm{torque}}$
& Regularizes the net body torque magnitude to encourage energy-efficient actuation.
& $\sum_i \|\boldsymbol{\tau}_i\|_1$ \\[3pt]
\hline
\end{tabular}
\end{table*}

\section{Experiments}
\label{sec:exps}
We evaluate four actuation configurations to isolate the role of Int-Ball2's drag-torque mechanism~\cite{Mitani2023IntBall2} and rotor-polarity structure in RL-based docking performance. For all configurations, the robot is initialized at varying, dynamically reachable poses inside the JEM volume. The details are included in Table \ref{tab:configurations}:

\begin{table}[t]
    \centering
    \caption{Propulsion-model and reward-shaping configurations used for
    ablation. 
    DT = drag-torque dynamics in physics; 
    Polarity = rotor spin pattern \cite{Mitani2023IntBall2}; 
    Penalty = drag-torque penalty $R_{drag}$.}
    \label{tab:configurations}
    \begin{tabular}{lccc}
        \toprule
        \textbf{Config} & \textbf{DT Dynamics} & \textbf{Polarity} & \textbf{Penalty} \\
        \midrule
        A (Baseline) & Disabled & Alternating & Enabled \\
        B            & Enabled  & Alternating & Enabled \\
        C            & Enabled  & Same (all +1) & Enabled \\
        D            & Enabled  & Alternating & Disabled \\
        \bottomrule
    \end{tabular}
\end{table}

Each configuration is trained with three seeds and evaluated in headless deterministic mode over 300 parallel environments (900{+} docking attempts per condition). A bounded observation noise is injected during all training runs, where uniform perturbations $\epsilon \sim \mathcal{U}(-\delta, \delta)$ are independently applied to the four slices of the 15-dimensional observation vector: position error ($\delta = 0.03$ m), 6D orientation representation ($\delta = 0.01$), linear velocity ($\delta = 0.03$ m/s), and angular velocity ($\delta = 0.03$ rad/s). This noise models sensor uncertainty but is not treated as an ablated experimental factor.

\subsection{Success Definition}
\label{sec:success_def}
We deem a docking attempt as successful if the agent achieves and maintains, for at least 5 consecutive timesteps, both the position and orientation errors below their respective thresholds as defined below:
\begin{itemize}
    \item a relative position error below $2\,\text{cm}$, and
    \item an orientation error below $2^{\circ}$
\end{itemize}
with respect to a set target pre-docking pose (i.e., the pose immediately prior to magnetic capture). Orientation error is computed from the trace of the relative rotation matrix reconstructed from the 6D continuous representation used in the observations~\cite{zhou2019continuity}. For each episode, evaluation metrics are taken at the timestep of minimum positional error close to the end of the episode, which offers a physically meaningful assessment of docking accuracy.

\subsection{Results}

\begin{table*}[t]
    \centering
    \caption{
        Docking performance across the four configurations described in Section~\ref{sec:exps}. Docking Success requires maintaining both position and orientation errors within threshold for at least five consecutive timesteps. Momentary Achievement reports the percentage of episodes that satisfy both thresholds at the evaluated docking point (timestep of minimum distance-to-dock, with orientation evaluated over a $\pm5$-step window).
    }
    \label{tab:docking_performance}
    \begin{tabular}{lcccccccc}
        \toprule
        \textbf{Config} &
        \makecell{\textbf{Final}\\\textbf{Pos Err (m)}} &
        \makecell{\textbf{Final}\\\textbf{Ori Err (deg)}} &
        \makecell{\textbf{\% Pos}\\\textbf{$<$ thresh}} &
        \makecell{\textbf{\% Ori}\\\textbf{$<$ thresh}} &
        \makecell{\textbf{\% Momentary}\\\textbf{Achievement}} &
        \makecell{\textbf{\% Stable}\\\textbf{Docking Success}} &
        \makecell{\textbf{\% Time}\\\textbf{Pos $<$ thresh}} &
        \makecell{\textbf{\% Time}\\\textbf{Ori $<$ thresh}} \\
        \midrule
        A & 0.009 & 3.90 & 96.4 & 38.6 & 37.6 & 65.6 & 6.7 & 4.6 \\
        B & 0.008 & 0.67 & 98.1 & \textbf{99.1} & \textbf{97.3} & 97.2 & \textbf{10.7} & 35.1 \\
        C & 0.128 & 7.03 & 3.0  & 14.4 & 0.8   & 0.1  & 0.0 & 2.1 \\
        D & \textbf{0.007} & \textbf{0.60} & \textbf{99.7} & 97.7 & \textbf{97.3} & \textbf{99.3} & 9.1 & \textbf{35.4} \\
        \bottomrule
    \end{tabular}
\end{table*}

The study shows a clear effect of physically accurate drag-torque modeling and polarity structure on the modelled docking performance. Including drag-torque dynamics with the intended alternating CW/CCW polarity yields consistently stable final alignment and the most reliable docking behavior, which shows that the additional torque channel is both
useful and effectively exploited by the learned controller. Removing the drag-torque penalty (Config~D) leads to an ideally sparse, however, unstructured actuation pattern (see Figure \ref{fig:prop_usage}), where only a subset of propellers is heavily used while others remain largely inactive. This indicates that the penalty is necessary to regularize thrust allocation and maintain the balanced, polarity-consistent propeller usage.

In contrast, enforcing identical spin polarity across all propellers severely degrades fine attitude control, leading to poor convergence and frequent orientation failures. This directly reflects the physical importance of polarity symmetry for passive drag-torque cancellation in Int-Ball2. The thrust-only baseline achieves reasonable positional approach but struggles with precise orientation, highlighting that accurate modeling of the drag-torque mechanism materially improves the controllability and stability of the final docking phase.

\begin{figure}[t]
    \centering
    \includegraphics[width=\columnwidth]{figs/propeller_usage_heatmap.pdf}
    \caption{
    Mean per-propeller action magnitude $|u_i|$ over a
    $\pm 5$-timestep window around closest approach (“near docking”),
    averaged across all seeds and 300 environments for configurations A to D.
    }
    \label{fig:prop_usage}
\end{figure}

\section{Conclusion}
\label{sec:conclusion}

This work shows that physically accurate modeling of Int-Ball2's propeller drag-torque dynamics and polarity structure improves docking stability in constrained intravehicular environments, and that PPO can exploit these effects to achieve reliable 6-DoF docking inside the ISS JEM. The ablations highlight how symmetry and residual-torque regularization shape stable attitude control, underscoring the interaction between actuation physics and learned policies. Looking forward, safety-aware maneuvering of Int-Ball2 inside the JEM remains an open challenge, motivating extensions toward collision-aware navigation and constrained RL methods that enforce safety and orientation limits directly. Real-world experiments using our modular sim-to-real framework~\cite{elroboran} will further enable hardware validation and pave the way for deployable autonomy in next-generation intravehicular free-flyers.

\section*{Acknowledgments}

The Int-Ball2 visual model and the JEM module used in our simulator were adapted from the open-source \texttt{int-ball2\_isaac\_sim} package by SpaceData Inc.~\cite{spacedata_intball2_isaac_sim}. The dynamics, control, and reward logic are our own.


\bibliographystyle{plainnat}
\bibliography{references}

\end{document}